# Exploring and Reshaping the Weight Distribution in LLM

**Chunming Ye**[1*]   **Songzhou Li**[1]   **Xu Xu**[1]

1 Suzhou University

{winshepherd@163.com, songzhouli@ahszu.edu.cn, xuxu@ahszu.edu.cn }

## Abstract

The performance of Large Language Models is influenced by their characteristics such as architecture, model sizes, decoding methods and so on. Due to differences in structure or function, the weights in different layers of large models have varying distributions. Taking the decoder-only LLaMA model as the research object, this paper explores the correlations between different types of layers in terms of weights distribution and studies the potential impact of these correlations on LoRA training effectiveness. Firstly, the study reveals that in the model the cosine distances between weights of different layers manifest power-law distribution. We extract Query-projection, down-projection and other weight matrices from the self-attention layers and MLP layers, calculate the singular values of the matrices using singular value decomposition, and organize a certain number of singular values into matrices according to projection's type. By analyzing the probability distribution of the cosine distances between these matrices, it is found that the cosine distances values between them have distinct power-law distribution characteristics. Secondly, based on the results of distance calculations and analysis across different layers of model, a qualitative method is proposed to describe the distribution characteristics of different models. Next, to construct weights that align with the distribution characteristics, a data generator is designed using a combination of Gaussian process and Pareto distribution functions. The generator is used to simulate the generation of data that aligns with specific distribution characteristics. Finally, based on the aforementioned distribution characteristics and data generation method, the weights in LoRA initialization are reshaped for training. Experimental results indicate that, without altering the model structure or training process, this method achieves a certain improvement in the performance of LoRA training.

## 1. Introduction

Although the enormous number of weight parameters endows large models with the capability to tackle many complex tasks, full-parameter fine-tuning of these weights demanded substantial computational resources. LoRA (Hu et al., 2022) was a method based on low-rank decomposition, which decomposed the LLM weight matrix into two low-rank A and B matrices. By updating the weights only in A and B matrices, LoRA significantly reduced resource consumption during training and can enhanced the model performance in specific tasks. During recent years, LoRA had garnered significant attention across multiple research domains in LLMs, with related studies (Mao et al., 2025) covering LoRA applications, downstream tasks, cross-task generalization, efficiency-enhancing methods, and more. Since LoRA traded off performance for reduced training resource consumption, many studies had sought various methods to improve LoRA training such as modifying the neural-network architecture, adjusting the training procedure, and optimizing weights or gradients. Some studies (Tian et al., 2024) enhanced training effectiveness by modifying the structure of the A and B matrices or introducing federated learning (Guo et al., 2024). Others achieved better training outcomes by freezing (Zhang et al., 2023)(Liu et al., 2024) or pruning certain weights (Zhang et al., 2023). Some research focused on parameters adjustments during training, such as aligning the gradients of low-rank matrices products (Wang et al., 2024) or decomposing weights (Liu et al., 2024)  and independently training certain components to improve performance.

These research findings had improved LoRA training for large models to varying degrees. However, the original intent of LoRA was to sacrifice some performance in order to reduce complexity of training and minimize resource consumption. Employing overly complex methods may inflate the training steps and resource demands, thereby undermining the very purpose of LoRA. Inspired by recent research findings on the impact of model-specific weights, activation values, and training parameter initialization on model performance, this paper explores the characteristics of model parameters to analyze potential correlations between weights distribution and model performance. The aim is to propose a simple, novel initialization scheme that improves LoRA training without increasing resource consumption or altering the neural network structure.

---

[*] Corresponding author

Since the introduction of neural network model, researchers had continuously explored the relationships between neural network structure, parameter characteristics, and models' performance. Subsequent research revealed that the structural composition of the network significantly impacts model performance. Various neural network models with different structures were designed and applied to diverse scenarios, including Feedforward Neural Network, Convolutional Neural Networks, Recurrent Neural Network, Long Short-Term Memory, Bidirectional Encoder Representation from Transformers, and various types of large language models (LLMs). These refinements on architecture have substantially improved model performance. Concurrently, research about the characteristics of weights distributions had never ceased. In early studies Gaussian distribution (Giryes et al., 2016)，uniform distribution, Marchenko-Pastur distribution and Student's t distribution were used to describe weights in model. However, some research (Bellido et al.,1993) thought these distributions are not fit many models. Some research (Barsbey et al.,2021) found that many neural networks had heavy-tails in weights after large step-size/batch-size ratios training. The other studies (Ahmed et al., 2018) suggested that the weights of NN models after training conform to the T-location scale statistical distribution rather than the normal distribution. Other distribution models, such as multivariate α-stable distribution (Li et al., 2024), were also used in modeling and analyzing model parameters.

Although changing and optimizing the architecture is considered as the main direction of research for model performance, studies (Narkhede et al., 2022) on analyzing how the performance is affected by distribution characteristics of weight, gradient and other parameters have never stopped. The most common method (Lee et al., 2023) was modeling neural networks and perceptron weights with gaussian distribution. Early studies (Go et al., 1999) analyzed the correlation between neural network weight distribution and model training. The largest eigenvalues of the covariance matrices were used for clustering to provide good initial weight points and to guide the training process. Some studies (Zhong et al., 2022) suggested memory capacity and generalization performance of perceptron is constraint on the distribution of its weights. A teacher-student framework with student perceptron weights was developed to mimic unknown teacher perceptron weight with lognormal and Gaussian distribution. In convolutional neural networks, the weights of the model were also considered to follow specific distributions (Huang et al., 2021). Through experiments it was found that well-trained convolutional filters approximately follow a Gaussian-alike distribution. With increasing research on weight distributions for analyzing and modeling, researchers have applied weight distributions to parameter initialization in model training. Some distributions(Thimm et al., 1995) such as uniform distribution, normal distribution and uniform over the intervals were employed for initial random weight distribution in neural network. In addition to these basic distributions, some non-linear activation functions (Kumar S.K., 2017) were also proposed to initialize weigh and implied strategy for the Rectified Linear Unit better than Xavier initialization. Other study (Wong et al., 2024) found Xavier initialization with linear activation function and Nguyen-Widrow weight initialization with hyperbolic tangent sigmoid activation function are better than uniform random values for initialization.

With the emergence of LLMs, model structures had become more complex, the number of parameters had increased greatly, and research on parameters also increased accordingly. These studies focused on various characteristics of weights, such as algebraic calculations of weight values, distinctions between weight types(layers) and their importance, weight initialization, the relationship between the number of weights and their effectiveness, weight and quantization and so on. The idea of calculating the weight values is just as the method described in the article (Chronopoulou et al., 2023) that combining weights with text clustering and semantic similarity among several adapters. This weight-space-averaging method was thought that it consistently improved trained adapters performance to new domains. Research focused on weights in different layers (Yu et al., 2024) found that the down-projection weights were regarded as super weights in LLM. These weights had a greater impact on quantized model's quality than other layers weights. The super weights induced super activations had lasting effects throughout the entire model. Concerned with different importance of weights，other study (Oh et al., 2024) selected top-k magnitudes from weights in LLM and proposed a method for massive weights curriculum dropout. Inspired by the super weights methods, distribution shift weight was(Jiang et al., 2025) proposed to calculate from consistency-filtered dataset for LLM self-improvement. To initialize weights some method (Dodge et al., 2020) try to modify the random seeds that control weight initialization of the final classification layer and training data. This method led to substantial performance gains on GLUE benchmark. The study (Lin et al., 2024) about small-scale models proposed an activation-aware per-channel scaling method to help bridge the performance degradation on-device LLMs. The selected salient weights which were considered more important for LLMs' performance were retained the precision of these parameters as FP16 and the other unimportant weights were quantized. Even the weights discarded in quantization were discussed to improve LLMs' accuracy (Yu et al., 2025). To improve LoRA training effectiveness, other parameters such as singular components of the pre-trained weight matrices had attracted some research. Pissia (Meng et al., 2024) initialized the adapter with principal singular values and vectors to for fine-tuning, while keeping the residual components frozen. MiLoRA (Wang et al., 2024) initialized the adapter with the minor singular components and frozen principal singular components. NORM (Jiang et al., 2025) also decomposed the LoRA parameters into two parts called essential and noisy components with singular value decomposition and remove the redundant parts for updating LoRA parameters efficiently.

It can be inferred from the existing research that: (i) The importance of weights in different layers varies for LLM; (ii) The optimization methods to weight characteristics, including quantization, freezing certain weights, initialization and so on, can help to improve model training performance; (iii) The effects of these techniques are more pronounced for small language model. Building on the above ideas, this paper explores simple methods to adjust the weight of small models for LoRA fine-tuning improving. The idea is how to reshape the small model structurally similar to a larger model without altering the architecture or training process, but rather through parameter initialization, thereby enhancing LoRA training effectiveness.

The main contributions of this paper:

(1) We analyze the weight distribution characteristics of LLM, construct each layer's singular value matrix through singular value decomposition, calculate the cosine distance between vectors of singular value matrices, and find that the cosine distances manifest power-law distribution characteristic. This characteristic is present in various types of LLMs

(2) Based on the cosine distance distribution characteristics between different matrices, a qualitative method is proposed to describe the model weight, enabling a more intuitive comparison of distribution characteristics between different models.

(3) A novel method is designed to generate data with their cosine distances conforming to specific distribution. This method is applied to initialize the tensor of the A and B matrix in LoRA according to the distribution characteristics to the reference model. The effectiveness of this method has been verified through training several models and testing on several benchmarks.

## 2．Analysis on distributions of weights

### 2.1 Reconstruct weights matrices

Taking the Decoder-only LLaMA model as the research object, we analyze the parameter characteristics of the model. LLM has too many weights and parameters. Even in 1B or 3B models, each layer contains hundreds of thousands or millions of weights. It is not aa easy work to compute and analyze all the weights. Refer to the LoRA approach, we concern with the weights matrices in LoRA for fine-tuning, including Query, Key, Value, Output projection matrices of weights in self-attention layers, and gate, up, down projection matrices of weights in feedforward network. We decompose the weight matrices using singular value decomposition (SVD) (Klema et al. 1980) to reduce computational complexity at the cost of some accuracy.

Let the weight matrix in any given layer be represented by matrix $X$. Performing singular value decomposition on $X$ yields the corresponding $U$, s, and $V$. The specifics are as follows.

Let $X \in R^{m \times n}$, there exist $U \in R^{m \times m}$, $V \in R^{n \times n}$ and $s \in R^{m \times n}$, then

$$X = UsV$$

where $s = diag(\sigma_1, \sigma_2 ..., \sigma_r) \in R^{m \times n}$. $r$ is the rank of $X$, $\sigma_i > 0$ are called singular values of $X$.

Taking the LLaMA-3.2-1B-Instruct model as an example, there are 16 layers in the model which can be used in LoRA training. Each layers contains self-attention block with Q-proj, K-proj, V-proj, and O-proj, as well as MLP block with gate-proj, up-proj, and down-proj. Each projection matrix in the layer is composed of the tensors of weights. We extract first $r$ singular values from the diagonal matrix $s$ to form a vector representing the features of matrix $X$. The vector composed of $r$ singular values is abbreviated as $SV_r$. If the model has a total of $L$ blocks, the corresponding weight features for all block in layers are denoted as $SV_1, SV_2 ..., SV_i, i < L$. The LLaMA 3.2-1B model has 16 layers. Each layer contains 7 different blocks: Q, K, V, O, gate, up, and down. Then 112 distinct $SV$r are extracted through singular value decomposition from the model weights' projection matrix.

The reason for selecting only $r$ values from diagonal matrix $s$ is that the weight matrices of each layer in the model have different dimensions. For example, in LLaMA-3.2-1B model, its $Q\_proj \in R^{2048 \times 2048}$, and $K\_proj \in R^{2048 \times 512}$, $down\_proj \in R^{8192 \times 2048}$. So the numbers of non-zero values obtained from each diagonal matrix $s$ are not the same. For the convenience of calculation and analysis comparison, referring to the concept of rank in LoRA, we select only the top-$r$ values from matrix $s$ to construct a new vector to represent the characteristics for weight matrix in the block.

A layer in LLaMA model contains Q, K, V, O blocks in self-attention sublayer and gate, up, and down blocks in MLP sublayer. Many studies take all the block of each layer as a whole (Zhang et al., 2023). We think after training the weights and their relationship remain stable. So, the different blocks such as Q, K, V and other projection matrix perform different functions during computation and inference. We also think the same type of blocks in different layers perform similar functions. In other words, the weight distributions of projection matrices with the same type are more similar. We combine $SV$ with the same kind of blocks in sequence into a matrix, call it the Matrix of $SV$, abbreviated as $MSV$. For the LLaMA3.2-1B model, we construct a total of 7 MSV corresponding to different blocks in all layers, which are $MSV_Q$, $MSV_K$, $MSV_V$, $MSV_O$, $MSV_{gate}$, $MSV_{up}$ and $MSV_{down}$. Each $MSV$ contains 16 $SV$, which composed of the singular values from the corresponding type of weight projection matrix in the 16 layers of the model. The purpose of constructing $MSV$ is to facilitate the analysis and calculation of weight matrices with the same function. The process of building MSV is shown in Figure 1.

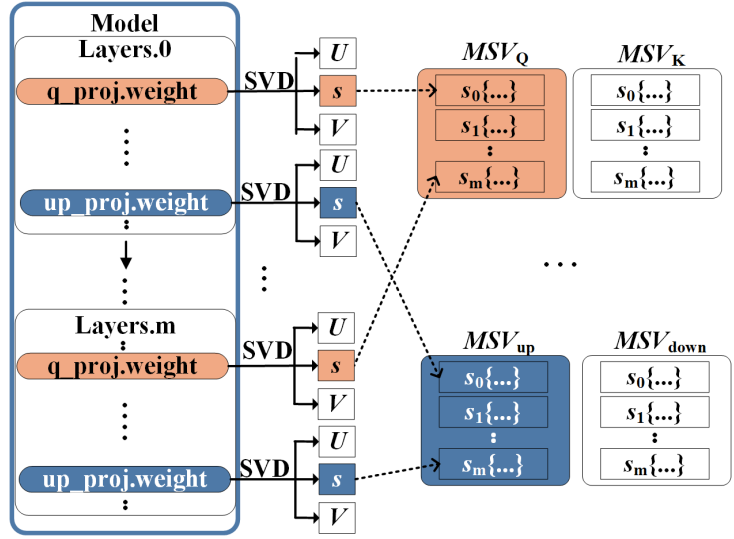

**Figure 1:** Decompose singular values from weight matrices in each layers, extract rank *r* singular values and build *MSV* for each kind of block

Unlike previous studies that directly analyzed weight projection matrices such as Q and K or their singular values, we analyze the *MSV* composed of singular values extracted from blocks in each layer. To measure the correlation between different *MSV*, we calculate the cosine distance between *SV* and performed statistical analysis. The formula (1) for calculating the cosine distance, denoted as $DSV_{x-y}$ which means the cosine distance between two vectors x and y.

$$DSV_{x-y} = \cos(x, y) = \frac{x \cdot y}{|x| \cdot |y|} = \frac{\sum_{i=1}^{n} x_i y_i}{\sqrt{\sum_{i=1}^{n} x_i^2} \sqrt{\sum_{i=1}^{n} y_i^2}} \quad (1)$$

We construct *MSV* of LLaMA3.2-1B-Instruct model, and plot the heatmap of cosine distances between *SV* in these *MSV*, as shown in Figure 2(a). For clearer display in this figure, the actual data is Cosine Similarity values, denoted as $\cos\_sim_{xy}$ as shown in the formula (2). Data in this figure are arranged according to original layers order in model. This graph shows that the cosine similarity between *MSV* do not show any obvious patterns for weight distribution. Some studies (Thamm et al., 2022) had done analysis on weight distribution and argued that the singular values of the trained model weights cannot be characterized by a tail index such as a power law type. But in this paper, we analyze the probability distribution of the cosine distances between *MSV*'s *SV* instead of using tensor of weight or singular values directly.

$$\cos\_sim_{xy} = 1 - DSV_{x-y} \quad (2)$$

### 2.2 Distribution characteristic of DSV

We calculate the distance *DSV* between all *SV* of LLaMA3.2-1B model and plot the probability distribution as shown in Figure 2(b), where the rank of *SV* is 16. The results show that the probability distribution exhibits power-law distribution pattern for the *DSV* between all *SV*. The we extract *SV* from LLaMA-3.2-3B-Instruct and LLaMA-3.2-8B-Instruct models and calculate *DSV* with *r* = 16. The probability distribution is shown in Figure 2(b). In result data all 0 value has been removed. The figure indicates that the *DSV* between all *SV* in both models also exhibit power-law distribution. Distances are calculated following the same steps with rank = 256. And the *DSV* probability distribution still exhibits power-law distribution, as shown in Figure 3(a). We selected the OLMo-1B, OLMo-7B, Qwen2-1.5B, and Qwen2-7B models for comparative analysis, setting rank =16, 64, and 256 respectively. The probability distribution of *DSV* distances is shown in Figure10 in the Appendix. As the model and rank values vary, the shapes of the probability distribution plots differ, yet all retain a clearly visible power-law distribution characteristics.

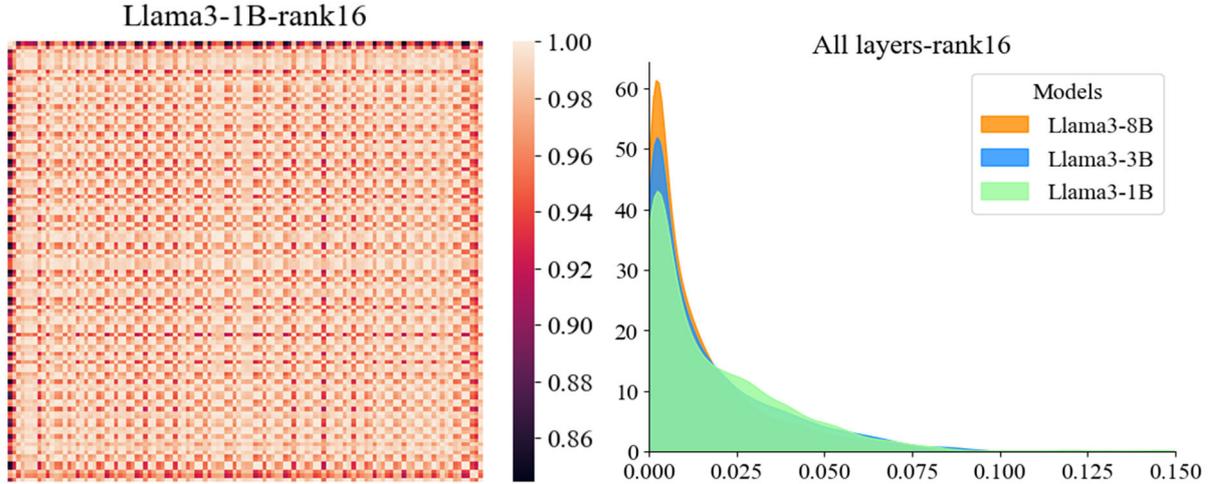

**Figure 2:** (a) Heatmap of distances between $SV$  (b) PDF of distances between $SV_{16}$ for LLaMA models

Set rank=16 and remove all 0 value from the distances (which is the distance value between SV and itself). The Python library scipy.stats.genpareto is used to fit the distance values to a Pareto distribution. The alpha values in the fitting results are shown in Table 1.

**Table 1:** Fitted Alpha values of Pareto distribution for models

| Models | Alpha value |
|---|---|
| LLaMA-3.2-1B-Instruct | 1.78 |
| LLaMA-3.2-3B-Instruct | 2.86 |
| LLaMA-3.2-8B-Instruct | 2.9 |
| OLMo-1B | 2.18 |
| OLMo-7B | 1.94 |
| Qwen2.5-1.5B- Instruct | 1.42 |
| Qwen2.5-7B- Instruct | 1.05 |

By analyzing the $DSV$ values between all $SV$, we find the principle that the cosine distances between different layers in model manifest power-law distribution characteristics. However, our goal is to discover the unique characteristics of different models in terms of weight distribution. Just a power-law distribution alone is not enough to identify the potential differences between models and between blocks. We make an assumption that since blocks such as Query, gate, and down have different functionalities, the cosine distance distributions between these blocks' $SV$ may vary, with some of which potentially deviating from the power-law distribution characteristics. Based on this assumption, we group $SV$ of the same block type together as a single entity, i.e., treating an $MSV$ as a single entity. Then we analyze the correlations in distance between different $MSV$. For the aforementioned LLaMA3.2-1B model, 7 $MSV$ are constructed, with each $MSV$ consisting of 16 $SV$. For example, $MSV_Q$ is composed of singular values extracted from Q-proj in 16 layers. When setting rank = 16, the $MSV_Q$ consists of 16 $SV$ with each length is 16, as $MSV_Q \in R^{16 \times 16}$. If let rank = 16, then $MSV_Q \in R^{16 \times 256}$. We calculate and analyze the $DSV$ distances between different $MSV$, such as the distance between $MSV_Q$ and $MSV_{\text{gate}}$, denoted as $DSV_{\text{Q-gate}}$. After calculation and plot, it is found that some $DSV$ still have power-law distribution. As shown in Figure 3(b), $DSV_{\text{Q-K}}$, $DSV_{\text{Q-gate}}$, and $DSV_{\text{V-up}}$ have power-law distribution characteristics. However, the others like distribution of $DSV_{\text{V-gate}}$ is clearly not power-law distribution but rather resembles a normal distribution. This confirms our assumption that not all distance distributions between $MSV$ follow power-law distribution, and the influencing factors may be related to model architecture, the functions of different layers or blocks, the number of weights, etc. In this paper we do not delve into the underlying causes but instead explore distribution characteristic and attempt to apply it to improve LoRA training.

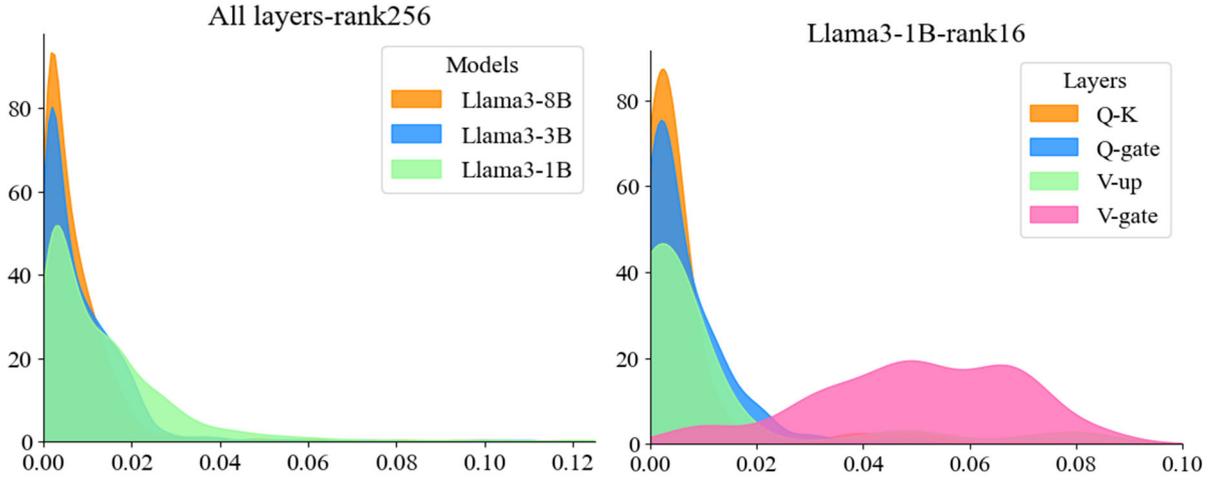

**Figure 3**: (a) power-law distribution for rank=256   (b) non-power-law distributions between different *MSV*

We analyze more *DSV* distribution in LLaMA-3.2-1B with rank = 16, and find that apart from $DSV_{\text{V-gate}}$, there are others *DSV* still exhibited non-power-law distribution characteristics, as shown in Figure 4, Figure 4(a) indicates that the distributions of $DSV_{\text{Q-up}}$ and $DSV_{\text{V-gate}}$ lie between power-law and normal distributions, while Figure 4(b) shows that distributions of $DSV_{\text{gate-up}}$ and $DSV_{\text{gate-down}}$ are more inclined toward a normal distribution. The analysis OLMo2 and Qwen models validated these, as shown in Appendix Figure 11, where non-power-law distribution characteristics of *DSV* are also observed between certain *MSV* in these models.

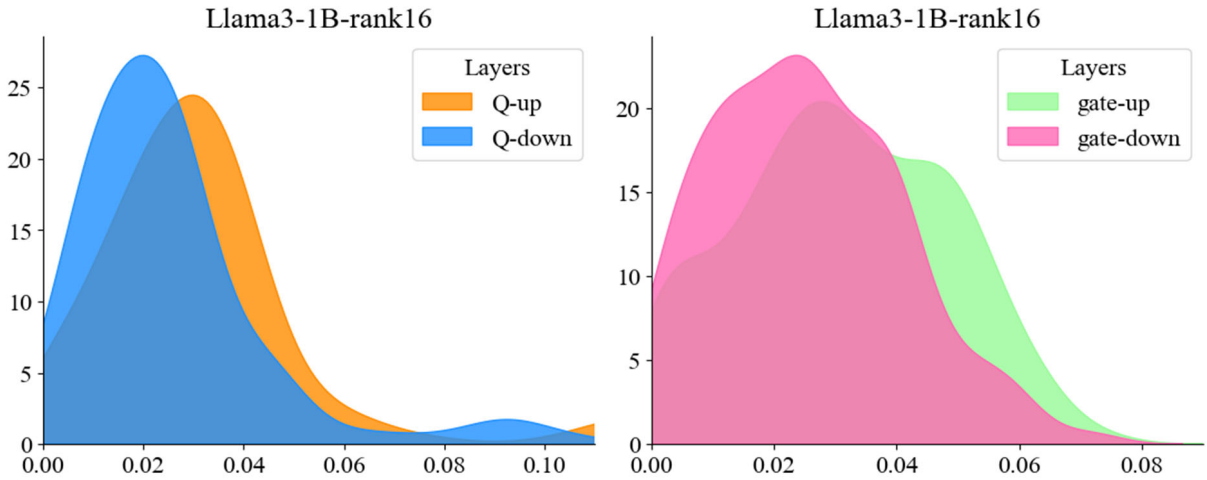

**Figure 4:** (a) Distributions of Q,up,down *MSV*   (b) Distributions of gate,up,down *MSV*

The above probability distribution plot has undergone some fitting during its creation. We plot some of these distribution data in polar coordinates, as shown in Figure 5. In these plots the distances are normalized to 360 degrees. The data of distributions for $DSV_{\text{Q-up}}$ and $DSV_{\text{gate-down}}$ come from LLaMA3-1B, LLaMA3-8B models. The distribution of values across different intervals appears more clearly in the figure, with distinct differences between power-law distribution and non-power-law distribution. Polar coordinate distributions for other models are shown in Appendix Figure 12.

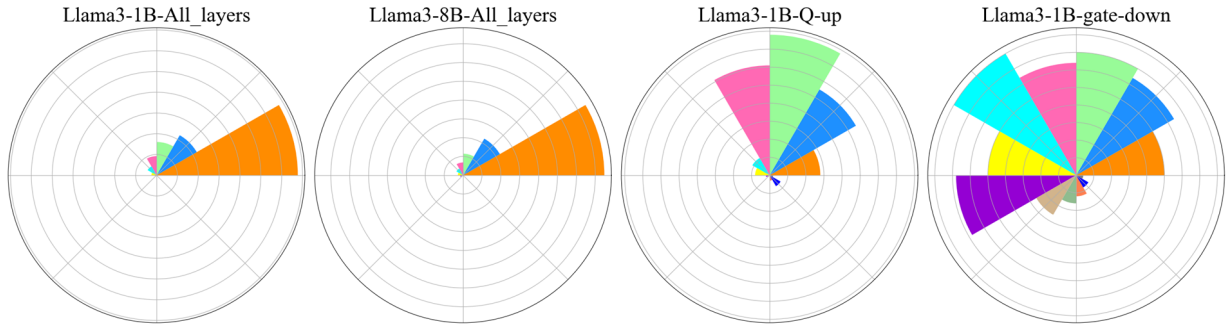

**Figure 5:** Distribution characteristics of LLaMA models on polar coordinates

The plots of same analysis on the smaller models SmolLM2-135M and SmolLM2-1.7B are shown in Appendix Figure 13. The results indicate that the power-law distribution characteristics of *DSV* still exists in smaller models. In SmolLM2-1.7B, there are also some non-power-law distribution characteristics of *DSV*.

After analyzing distributions for model weights of LLaMA3.2-1B/3B/8B, OLMo2-1B/3B, Qwen2-1.5B/7B, and SmolLM2-135M/1.7B, the following conclusions are drawn.

(1) The power-law distribution characteristics of the cosine distance between weights matrices singular values are widely present across many different LLMs. This phenomenon is independent of factors such as model type, model scale, the rank of the singular values of the weight matrices, and so on. We speculate that the cause is as follows: the weight matrix of each layer or block comprises an enormous set of values, many of which are numerically close to one another. Their mutual differences become tiny relative to the total count of weights in matrices. Consequently, the singular-value vectors (*SV*) derived from the weight matrices exhibit pairwise cosine distances similarities, with most distance values being concentrated near 0. This concentration causes the overall probability distribution of the distances to display a pronounced power-law character.

(2) Some distributions of *DSV* do not conform to power-law distribution characteristics and are similarly widespread across many models. The distribution plots exhibit significant differences, approximating power-law distributions, lying between power-law and normal distributions, or approaching normal distributions. The difference between gate-proj, down-proj and the other projections is particularly significant. This could be influenced by factors such as the functionality of the weight matrices themselves, the model's architecture, the number of model weights, even the training data and so on.

(3) Power-law distribution characteristics are relatively common, but non-power-law distribution characteristics in specific layers or blocks vary across different models. This provides the possibility of characterizing different models by extracting distribution features.

Based on the above research, we further hypothesize that models with more weights may perform better in inference tasks, as their weight distributions may be more reasonable. As one of the network architecture features, distribution characteristics may have a certain impact on model performance. If we adjust the weight distributions of smaller models to align with those of larger models, this may improve the performance of smaller models. Therefore, we will explore how to describe the distribution characteristics of models and how to generate weights that align with the corresponding larger model distribution characteristics.

## 3. Reshape Distribution

### 3.1 Characteristic of model's distribution

In the above study, *DSV* distribution characteristics are discussed and attempted to distinguish models. We propose a qualitative method to describe models. Through several rounds of observation and comparison, the characteristic of model would be determined according to correlation between *MSV*'s distributions. The explanation is that if the *DSV* between two *MSV* manifests power-law distribution, it indicates the weight tensors in these two *MSV* are very close. Otherwise, one of them may be closer with other *MSV*.

The first step of the method is selecting a particular *MSV* from the model as the first Referenced Weight, denoted as $RW_1$. The default Referenced Weight is $MSV_Q$ extracted from Q-proj in all layers. The next step is calculating *DSV* and analyzing the distribution relationship between $RW_1$ and other *MSV*. If $DSV_{Q-x}$ (x stands for other projection) has power-law distribution, $MSV_x$ is selected as a member weight of $RW_1$. Otherwise $MSV_x$ would not be member of $RW_1$ and the next *MSV* would be analyzed. This process continues until $MSV_Q$ has been compared with all other *MSV*. Among the non-member weights, another *MSV* is selected as the second Referenced weight ($RW_2$), and the comparison process is repeated until each referenced weight and its corresponding member weights manifesting power-law distribution characteristics are obtained.

Taking the LLaMA3.2-8B-Instruct model as an example, the *DSV* distribution of $MSV_Q$ and other *MSV* is shown in Figure 6(a). The figure shows that $DSV_{Q-k}$, $DSV_{Q-gate}$, and $DSV_{Q-O}$ exhibit power-law distribution characteristics, while $DSV_{Q-V}$, $DSV_{Q-up}$, $DSV_{Q-down}$ exhibit non-power-law characteristics. Then we conducted second comparison of $MSV_v$, $MSV_{up}$, and $MSV_{down}$. Through experimentation we selected $MSV_{up}$ as the second referenced weight, calculated the *DSV* between $MSV_{up}$ with other *MSV*. The results are shown in Figure 6(b). These plots indicate that $DSV_{up-v}$ and $DSV_{up-down}$ exhibit power-law distribution characteristics. The final analysis results for LLaMA3-8B are as follows: all *MSV*s are divided into two groups. The first group uses $MSV_Q$ as the referenced weight, and the *DSV* between $MSV_Q$ and $MSV_K$, $MSV_O$, and $MSV_{gate}$ exhibits power-law distribution. The second group uses $MSV_{up}$ as the referenced weight, and the *DSV* between $MSV_{up}$, $MSV_V$ and $MSV_{down}$ exhibit power-law distribution. The referenced weights and their members are shown in Table 2.

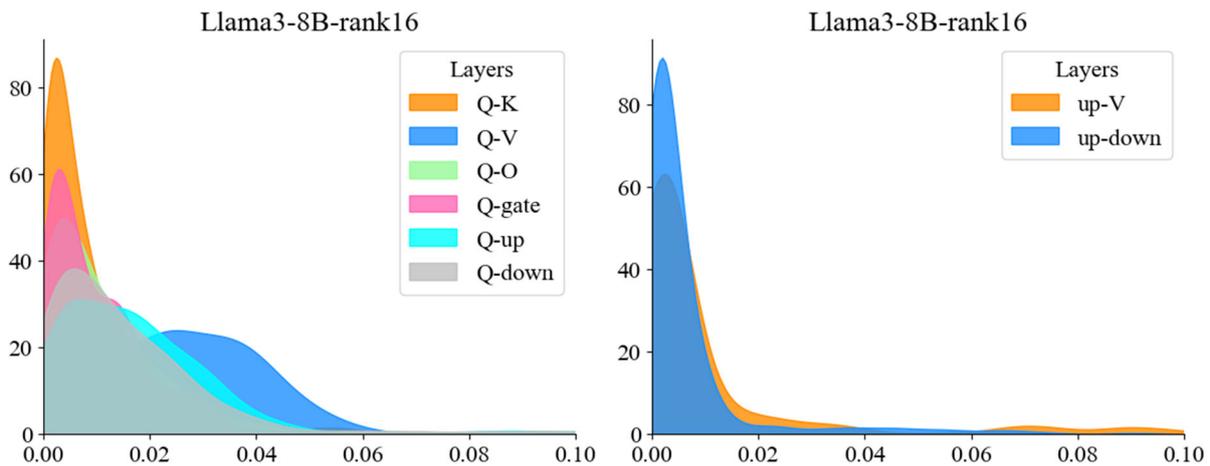

**Figure 6:** (a) Distributions of Q with other MSV    (b) Distributions of up with V, down MSV

Table 2: Characteristic of distribution for LLaMA3-8B

| No. | Referenced Weight | Member | | |
|---|---|---|---|---|
| 1 | Q-proj | K-proj | O-proj | gate-proj |
| 2 | V-proj | up-proj | down-proj | |

### 3.2 Simulating distribution with randoms

Based on the qualitative method, a distribution generator is designed to generate data that conforms to the expected distribution characteristics, and is planned to be used to reshape the initialization weights on LoRA. Gaussian Process and Pareto distribution function are used for data generation. The Pareto distribution is used to simulate power-law characteristics. And it along with Gaussian Process is used to simulate other distribution characteristics. It must be pointed out that it is not the weights projection matrices or *MSV* parameters themselves that conform to those distributions, but rather the cosine distance (denoted as *DSV*) between the *SV* in the *MSV* conforming to those distributions. The goal of the distribution generator is to generate these projection matrices or *MSV*, and ensures the *DSV* between them conforms to the expected distributions.

These random functions are plan to generate random numbers and obtain the required weight values through superposition. Assuming two matrices *A* and *B*, the following design is proposed.

If the distance $DSV_{A-B}$ between matrices *A* and *B* approximates Gaussian distribution, indicating little correlation between *A* and *B*. Then the data in matrices *A* and *B* can both be generated from Gaussian distribution.

If the distance $DSV_{A-B}$ between matrices *A* and *B* approximates power-law distribution, indicating that the data in matrices *A* and *B* differ only slightly. Suppose the data in matrix *A* follow Gaussian distribution. Then we generate a number of random numbers whose count follows Pareto distribution. Add these data to matrix *A* and get the final matrix *B*. It is expected the distance between *A* and *B* follows power-law distribution.

According to the assumption, the following steps are used to simulate the generation of two random number matrices $A \in R^{m \times n}, B \in R^{m \times n}$

Step 1: Generate a one-dimensional array following a Gaussian distribution as template weight, denoted as $T = \{t_1, t_2 ..., t_n\}, t_i \sim N(\mu, \sigma^2)$. The length of template weight is ***n***

Step 2: Generate incremental data $\Delta W$ to modify the template weight. The length of $\Delta W$ is *p*. First generate a random number ***p***, where ***p*** < ***n*** and $p \sim Func()$. The *Func*() is a constant or random function. Random functions include Gaussian Process and Pareto distribution function. If the generated ***p*** value is greater than ***n***, it would be replaced with ***n*** directly.

Step 3: According to the generated ***p*** value, generate an array $\Delta W$ which conforms to Gaussian distribution. The length of $\Delta W$ is ***p***, denoted as $\Delta W = \{d_1, d_2 ..., d_p\}, d_i \sim N(\mu, \sigma^2)$.

Step 4: Add the template data *T* and incremental data $\Delta W$ to generate the final initialization weights ***v***. In this paper, the two arrays are summed together in random order.

Step 5: Repeat steps 2-4 a total of *m* times to generate the matrix $A = \{v_1, v_2 ..., v_n\}, A \in R^{m \times n}$.

Step 6: Repeat the above steps to generate another matrix *B*.

The above steps are illustrated in the middle part of Figure 7.

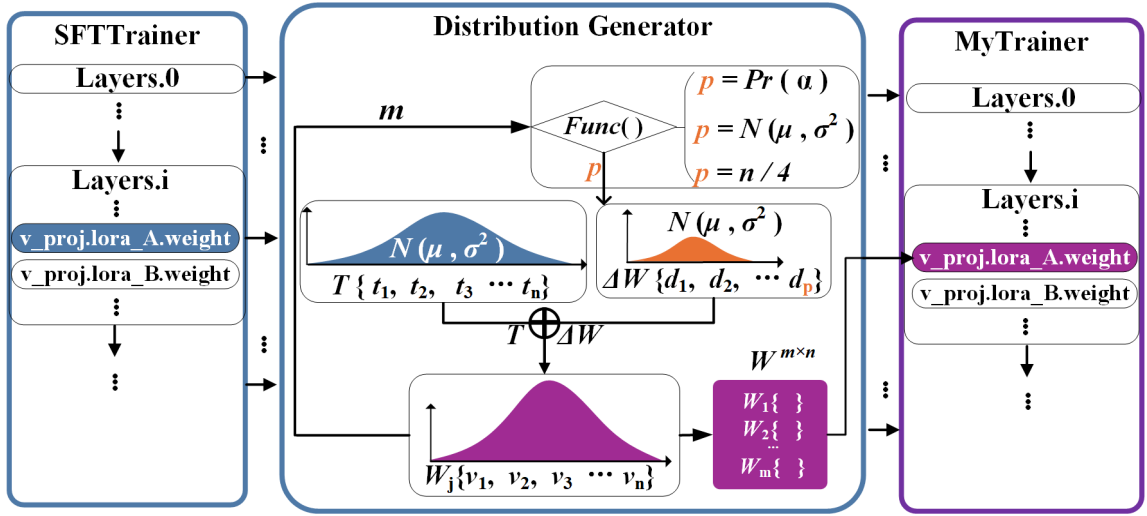

**Figure 7:** Reshape the LoRA weights with Distribution Generator

By modifying the type of *Func*() in the distribution generator (DG), two sets of experimental data were generated, each consisting of two matrices *A* and *B*. In the first set of experiments, *Func*() is Gaussian Process function. The data in matrices *A* and *B* are generated by adding new Gaussian distribution data to the original Gaussian distribution template data. The data in matrices *A* and *B* are labeled as Gaussian+Gaussian. The distance between *A* and *B* is labeled as Gaussian between Gaussian.

In the second set of experiments, *Func*() is changed to the Pareto distribution function to generate random numbers ***p***. This set of data is based on the original template data that follows a Gaussian distribution, with second group of new Gaussian distribution data random numbers added to it. The number of second group is p which follows Pareto distribution. This data is labeled as Gaussian+Pareto, and the distance between *A* and *B* is labeled as Gaussian between Pareto.

The result is plotted in Figure 8 . Figure 8(a) represents the *DSV* distributions between two sets of data in *A* and *B*. Figure 8(b) represents the distribution of the data themselves in *A* and *B*.

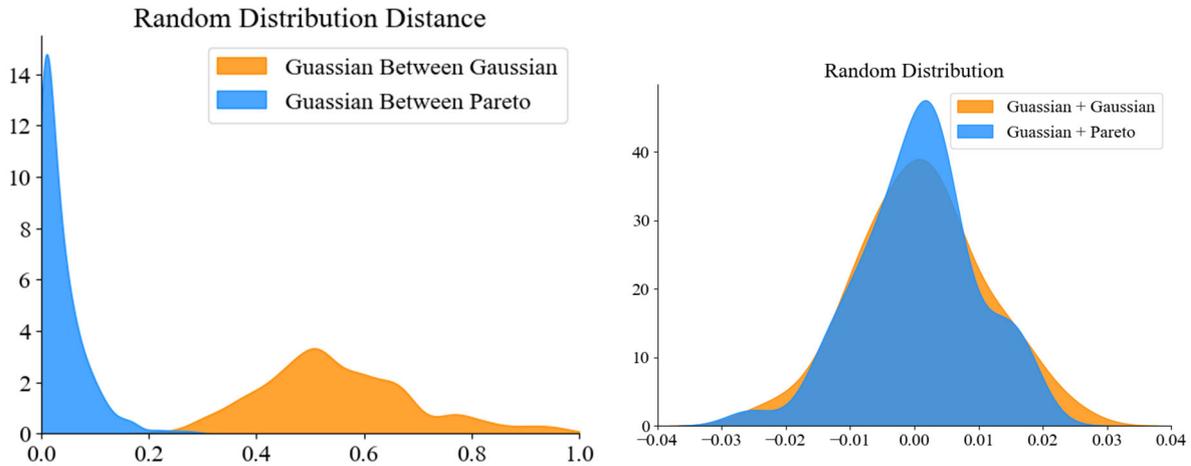

**Figure 8:** (a) Distribution of DSV from data generated by DG      (b) Distribution of data generated by DG

It can be seen from Figure 8(b) that the data constructed by distribution generator exhibits characteristics of approximate normal distribution regardless of the type of *Func*(). However, the distribution of *DSV* varies with the type of *Func*(). When *Func*() is a constant or Gaussian distribution, the *DSV* between *A* and *B* approaches normal distribution. When *Func*() is Pareto distribution function, the *DSV* between *A* and *B* exhibits power-law distribution pattern. This indicates that the distribution generator can be used to construct *A* and *B* matrices whose data and *DSV* between the data all conform to the expected distribution.

## 4．Experiment

### 4.1 Reshaping initial weights

In LoRA the default method is initializing low-rank matrices with Kaiming He or Gaussian distribution and zero values. In this paper combining the qualitative method and distribution generator, we reshape the weights in LoRA low-rank matrix *A* and matrix *B* during initialization. The experimental steps are as follows, with an overall schematic diagram shown in Figure 7.

(1) Select the pre-trained model X, and additionally select another model as the reference model M. The model X will initialize its low-rank matrices according to the distribution characteristic of model M.

(2) Extract the distribution characteristics of model M using the qualitative method.

(3) Reshape the LoRA low-rank matrix *A* and matrix *B* of model X according to the distribution characteristics of model M.

(4) Train LoRA model for X and evaluate with standard datasets.

The selection of reference model is based on two assumptions: First, it is assumed that the weight distribution of the current model to be trained is reasonable enough to be used as the reference model for reshaping LoRA low-rank matrices. Second, it is assumed that the larger model with more weights is more reasonable. So,a larger model is selected as the reference model to for pre-trained model's weight reshaping in LoRA initilization.

## 4.2 Results and analysis

In experiment the dataset is alpaca-gpt4-data-en (Peng et al. 2023). The unsloth[1] is used for LoRA fine-tuning training. The lm-evaluation-harness[2] is employed to evaluate the quality of fine-tuned model. The experiment utilized two small-scale pre-trained models: SmolLM2-135M (Allal et al., 2025) and LLaMA3.2-1B-Instruct (Dubey et al., 2024). In addition to the models themselves, the reference models also include SmolLM2-1.7B and LLaMA3.2-8B-Instruct, which have more weights. The evaluation is conducted using three datasets: GPQA Diamond Zero-Shot, Arc_Challenge, and HellaSwag.

The distribution characteristics extracted by each model are shown in Table 3. The distribution characteristic of SmolLM2-135M model shows that the $DSV$ between $MSV_Q$ and all other $MSV$ follow power-law distribution. It means there is no available distribution characteristic in SmolLM2-135M, and therefore this model itself cannot be used as a reference model. In SmolLM2-1.7B model, the $DSV$ between $MSV_Q$ and all other $MSV$ approximately follow power-law distribution except for $MSV_O$. According to these characteristics the Q-proj is used as the template weight for all projection matrices except O-proj. The LLaMA3.2-1B model uses Q-proj and V-proj as template weights. In LLaMA3.2-3B model, the $DSV$ distributions between $MSV_Q$, $MSV_K$, and $MSV_{up}$ are non-poer-law. So, Q-proj, K-proj, up-proj serve as template weights. In LLaMA3.2-8B model, Q-proj serves as the template weight for K-proj, O-proj, and gate-proj. When analyzing V-proj, up-proj, and down-proj, it is found that V-proj and down-proj are both closer to up-proj, so up-proj is selected as the template weight. The distribution characteristics of each model are shown in Appendix Figures 15-16.

In experiment Gaussian process is used to generate random values to initialize the Q-proj of layer.0 as the template weight. The $Func()$ is set to Pareto function. All members replicate the data from the template weight. Then distribution generator is used to reshape and initialize the values for each layer. Finally, the lora_A and lora_B matrices are initialized and generated for all layers. The experimental parameters are shown in Appendix Table 5. The loss values during training are shown in Figure 9. The results show that the loss values stabilize and converge after about 100 steps. It has with no significant difference from the loss in normal LoRA fine-tuning training.

**Table 3:** Characteristic of distribution for models

| Model | Referenced Weight | Member | | | | | |
|---|---|---|---|---|---|---|---|
| SmolLM2-135M | Q-proj | K-proj | V-proj | O-proj | gate-proj | up- proj | down-proj |
| SmolLM2-1.7B | Q-proj | K-proj | V-proj | gate-proj | up- proj | down-proj | |
| | O-proj | --- | | | | | |
| LLaMA3.2-1B | Q-proj | K-proj | gate-proj | | | | |
| | V-proj | O-proj | up- proj | down-proj | | | |
| LLaMA3.2-3B | Q-proj | gate-proj | | | | | |
| | K-proj | O-proj | | | | | |
| | up- proj | V-proj | down-proj | | | | |
| LLaMA3.2-8B | Q-proj | K-proj | O-proj | gate-proj | | | |
| | up- proj | V-proj | down-proj | | | | |

The evaluation results are shown in Table 4, where the accuracy values are the |acc_norm values in lm-evaluation-harness. In this table, SmolLM135M-LoRA refers to the evaluation results obtained after performing normal LoRA training directly on SmolLM135M. SmolLM135M-(Smo-1.7B) refers to the results obtained by training after initializing and reshaping the lora_A and lora_B matrices of SmolLM135M's LoRA using SmolLM2-1.7B as the reference model. LLaMA3-1B-(Self) refers to the results obtained by training after initializing and reshaping the lora_A and lora_B matrices using LLaMA3-1B as the reference model.

---

[1] https://github.com/unslothai/unsloth
[2] https://github.com/EleutherAI/lm-evaluation-harness

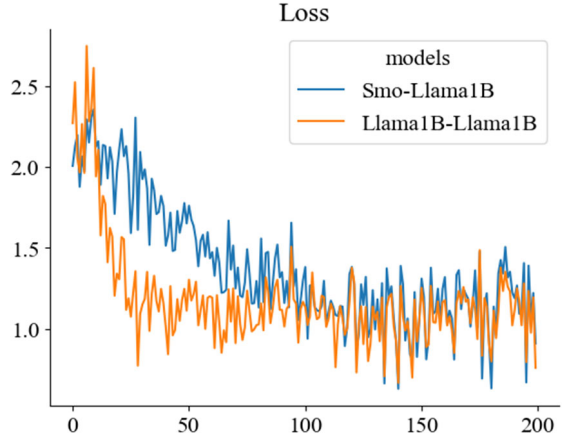

**Figure 9:** The loss of experiment. Smo-Llama1B means using LLaMA-1B as reference model to train SmolLM2-135M. Llama1B-Llama1B means using LLaMA-1B as reference model to train LLaMA-1B model.

**Table 4:** Result for benchmarks

| Model | GPQA Diamond Zero-Shot | Arc_Challenge | HellaSwag |
|---|---|---|---|
| SmolLM | 0.2222 | 0.2994 | 0.4299 |
| SmolLM135M-LoRA | 0.2222 | 0.2978 | 0.4386 |
| SmolLM135M-(Smo-1.7B) | 0.2172 | 0.2995 | 0.4364 |
| SmolLM135M-(LLaMA3-1B) | 0.2525 | 0.3046 | 0.4369 |
| SmolLM135M-(LLaMA3-3B) | 0.2525 | 0.2961 | 0.4374 |
| SmolLM135M-(LLaMA3-8B) | 0.2879 | 0.3038 | 0.4374 |
| LLaMA3-1B | 0.2727 | 0.3805 | 0.6077 |
| LLaMA3-1B-LoRA | 0.2727 | 0.3942 | 0.6117 |
| LLaMA3-1B-(Self) | 0.2929 | 0.3925 | 0.6118 |
| LLaMA3-1B-(LLaMA3-3B) | 0.2778 | 0.3891 | 0.6122 |
| LLaMA3-1B-(LLaMA3-8B) | 0.2929 | 0.3908 | 0.6121 |

The data shows that the reshaping distribution method can help to improve LoRA quality in most test. And specific effects vary depending on the reference model. First the SmolLM-135M is used as pre-trained model. Using the same series model SmolLM2-1.7B as reference model, the improvement on SmolLM-135M training is little. Its result only outperforms the test results of the pre-trained model. When using LLaMA3-1B/3B/8B as reference model, improvements on SmolLM-135M training are observed in all three datasets. Most evaluation results are better than both the pre-trained model results and normal LoRA training results. The improvement was particularly notable on the GPQA dataset test. The best results are achieved when reshaping and training using LLaMA3-8B as reference model. Second the LLaMA3-1B is used as pre-trained model. The test results of reshaping initialization method are better than original pre-trained model, and are slightly better than normal LoRA training results. The best test result is on GPQA dataset. When LLaMA3-8B is used as reference model for LLaMA3-1B, all tests have the best results. The experiment indicates by describing and reshaping during initialization, LoRA on small language model has better quality than pre-trained model and normal LoRA trained model. Selecting models with more weights as reference models may yield better performance. However, when using LLaMA3-3B as reference model, the performance is similar to that of using LLaMA3-1B as the reference model, which may be due to insufficiently precise description on distribution characteristics of LLaMA3-3B.

## 5. Conclusion

The original purpose of this paper is to find a simple way to improve the training effect of LoRA on large language model. Recent studies have revealed that different weights have different effects on LLM performance. Inspired by these studies and the idea of low-rank decomposition in LoRA, we try to analyze the characteristics of weight distribution in LLM. Using singular value decomposition, this paper extracts weight features from Query-projection, Key-projection, down-projection and other projection matrices across layers in LLMs. The singular values of these projection matrices are formed into new matrices. The analysis shows that the mostly cosine distances between these new matrices have typical power-law distribution characteristics, though some of them do not conform to power-law distribution. This paper proposes a qualitative method to determine these distribution characteristics and designs a distribution generator to simulate the generation of weight data which conform to expected distributions. The method are applied to initializing low-rank matrix $A$ and $B$ in LoRA to improve training. Tests on standard datasets show that this method can improve LoRA training performance.

## A. Implementation Details

Table 5: Configuration for Experiments

| Argument | Setting |
| --- | --- |
| OS & device | Ubuntu 24.04.1 LTS<br>NVIDIA GeForce RTX 4070 Laptop GPU 8G |
| per device train batch size | 2 |
| gradient accumulation steps | 2 |
| max_steps | 200 |
| dropout | 0.2 |
| learning rate | 1e-4 |
| warmup ratio | 10 |

## B. Distribution characteristics of models

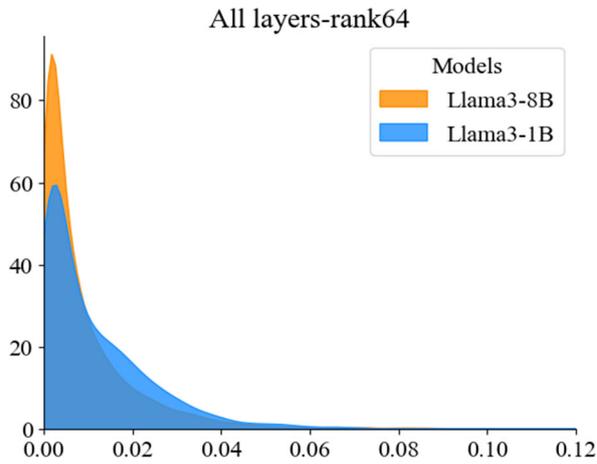

(a) Distributions of LLaMA models with rank=64

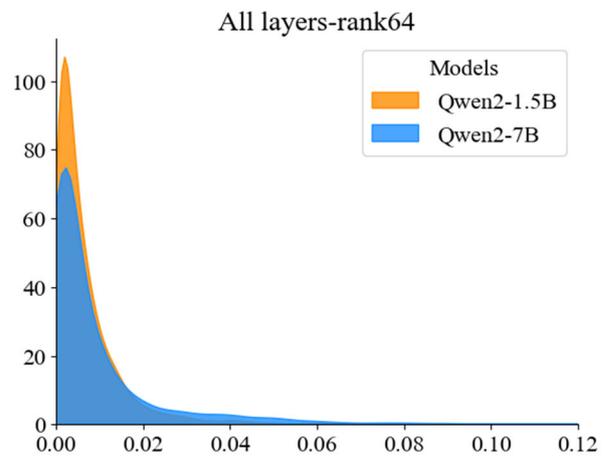

(b) Distributions of Qwen models with rank=64

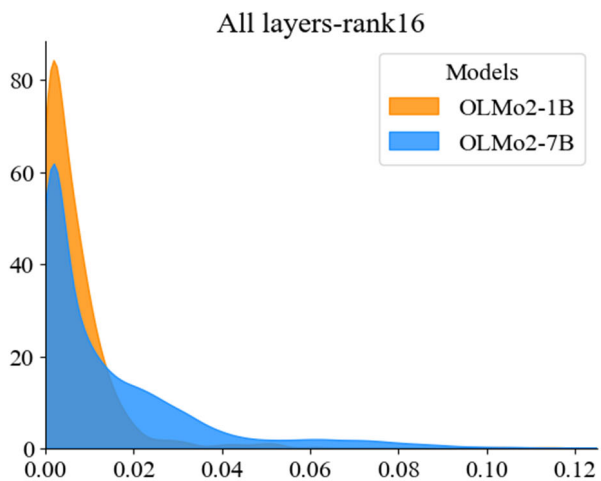

(c) Distributions of OLMo2 models with rank=16

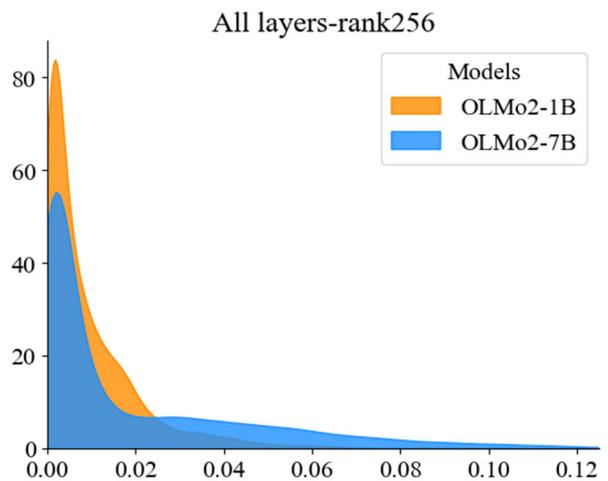

(d) Distributions of OLMo2 models with rank=256

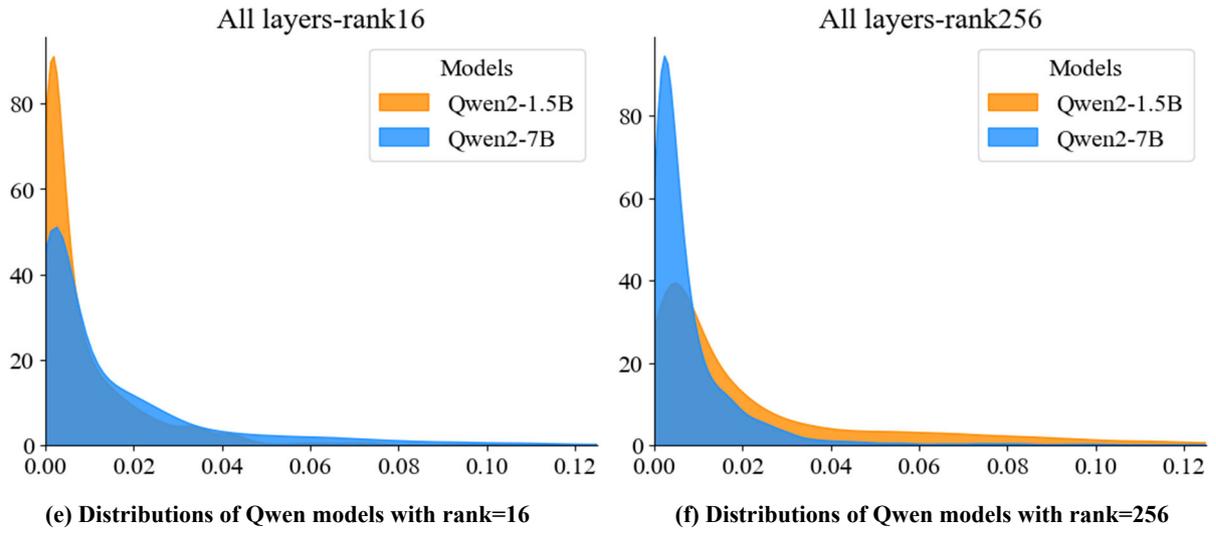

(e) Distributions of Qwen models with rank=16

(f) Distributions of Qwen models with rank=256

**Figure 10:** Distributions of LLaMA3, OLMo2, Qwen with different ranks.

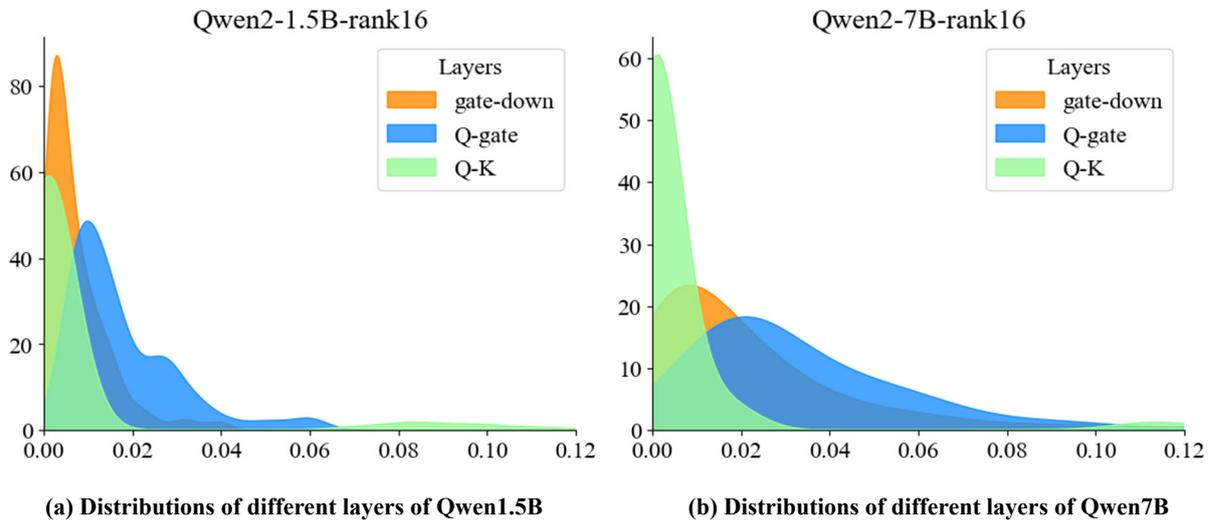

(a) Distributions of different layers of Qwen1.5B

(b) Distributions of different layers of Qwen7B

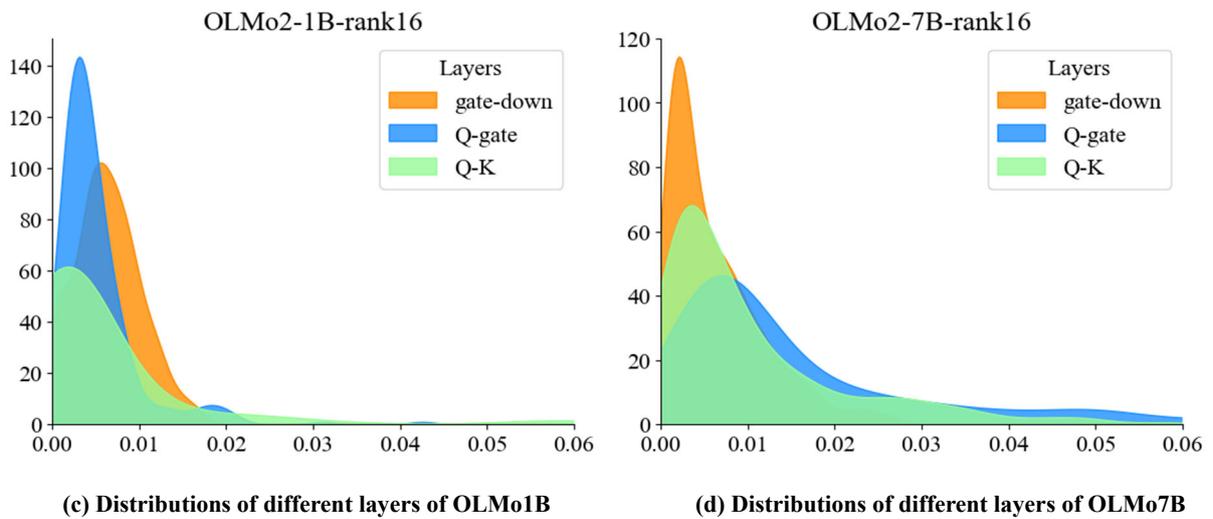

(c) Distributions of different layers of OLMo1B

(d) Distributions of different layers of OLMo7B

**Figure 11**: non-power-law distributions of different DSV in Qwen, OLMo models

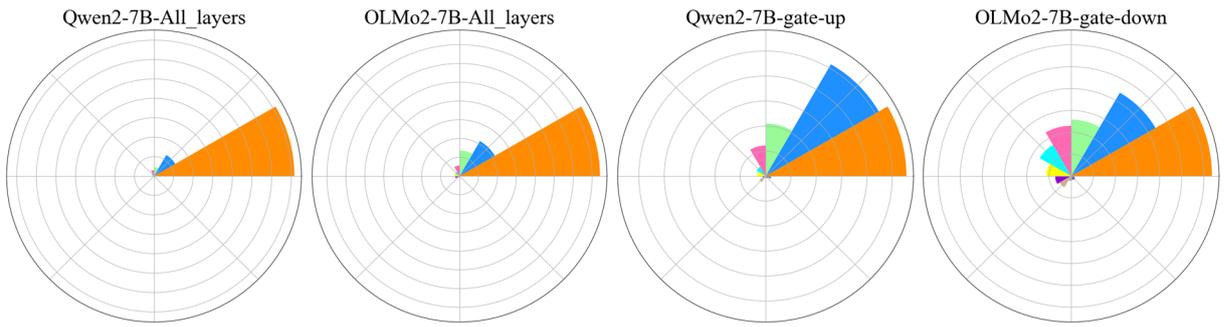

**Figure 12:** Distribution characteristics of OLMo,Qwen models on polar coordinates

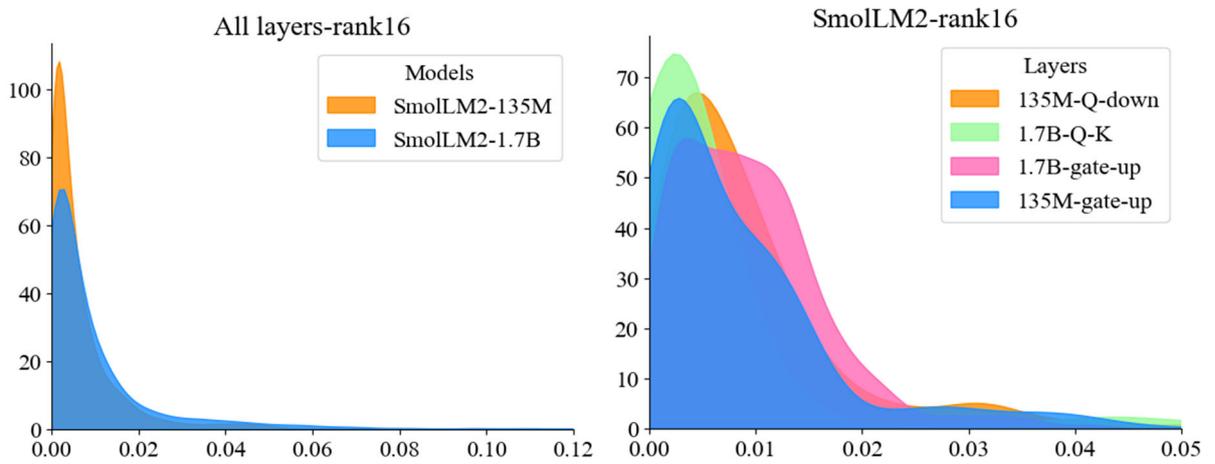

**Figure 13:** Distribution characteristics of SmolLM2 models

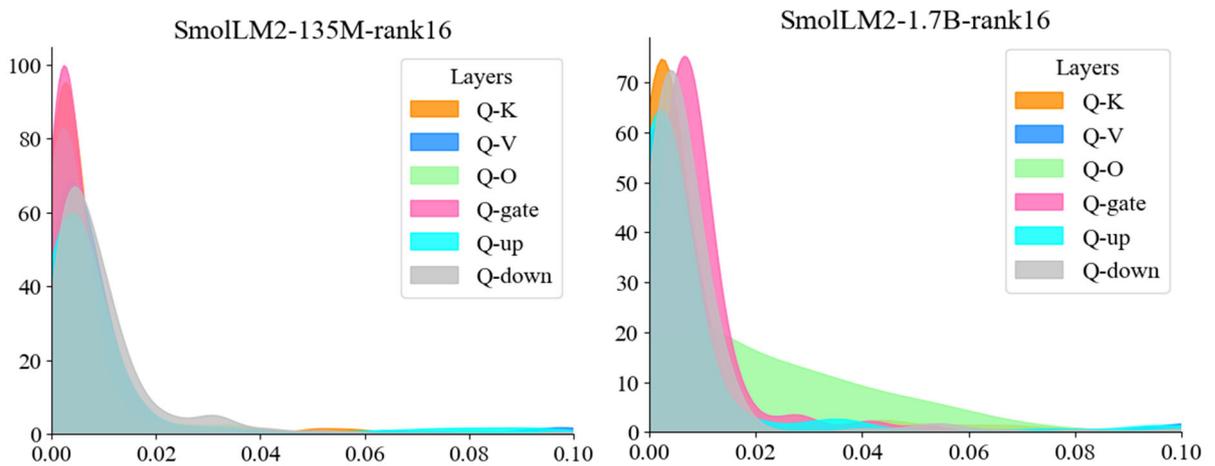

**Figure 14:** (a) Distributions of *MSV* for SmolLM2-135M  (b) Distributions of *MSV* for SmolLM2-1.7B

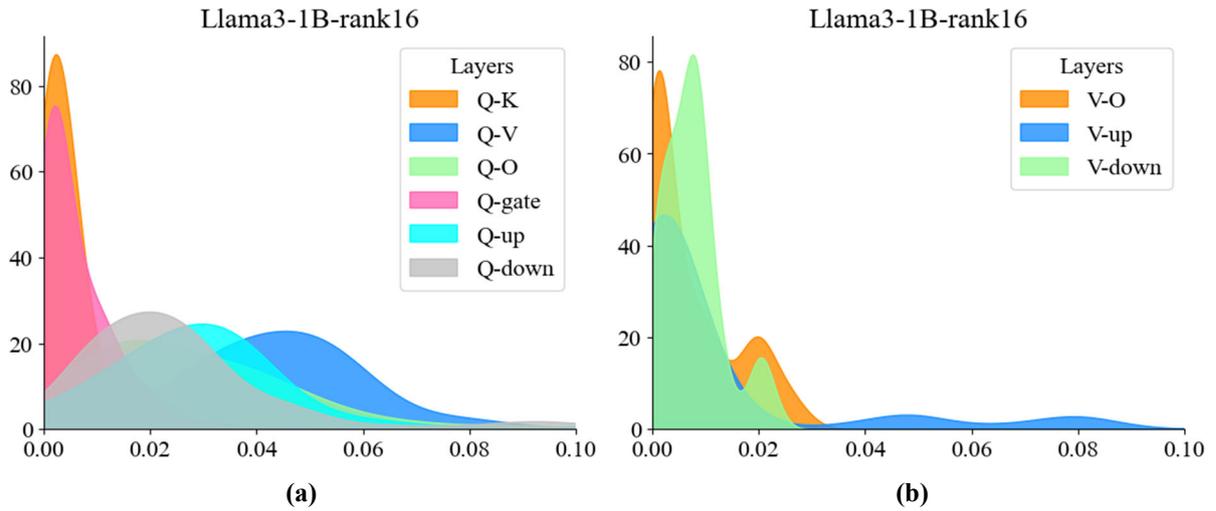

**Figure 15:** Distributions of *MSV* for LLaMA3-1B, (a) is distribution of $MSV_Q$ with other *MSV*, (b) is distribution of $MSV_V$ with O, up, down's *MSV*

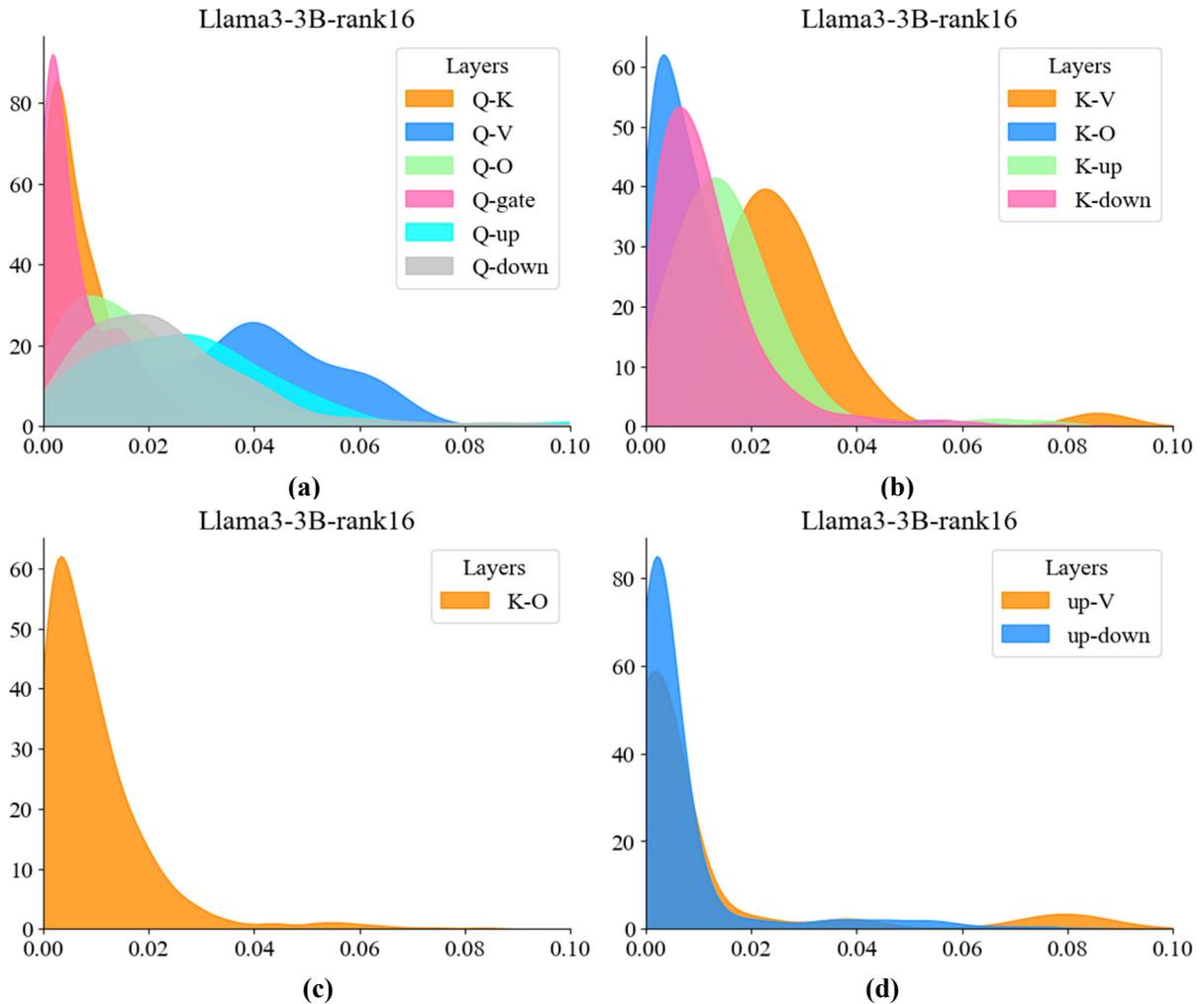

**Figure 16:** Distributions of *MSV* for LLaMA3-3B, (a) is distribution of $MSV_Q$ with other *MSV*, (b) is distribution of $MSV_K$ with V, O, up, down's *MSV*, (c) is distribution of $MSV_K$ with $MSV_O$, (d) is distribution of $MSV_{up}$ with V, down's *MSV*